\documentclass[]{spie}  

\usepackage{amsmath,amsfonts,amssymb}
\usepackage[ruled,vlined]{algorithm2e}
\usepackage{graphicx}
\usepackage{wrapfig}
\usepackage{caption}
\usepackage{subcaption}
\usepackage{booktabs}       
\usepackage[colorlinks=true, allcolors=blue]{hyperref}
\usepackage{array}
\usepackage{makecell}
\newcolumntype{P}[1]{>{\centering\arraybackslash}p{#1}}

\title{Adaptation of a deep learning malignancy model from full-field digital mammography to digital breast tomosynthesis}

\author[1]{Sadanand Singh}
\author[1]{Thomas Paul Matthews}
\author[1]{Meet Shah}
\author[1]{Brent Mombourquette}
\author[1]{Trevor Tsue}
\author[1]{Aaron Long}
\author[1]{Ranya Almohsen}
\author[1]{Stefano Pedemonte}
\author[1]{Jason Su}
\affil[1]{Whiterabbit AI, Inc., Santa Clara, CA, USA}

\authorinfo{Further author information: (Send correspondence to Whiterabbit.ai)\\Whiterabbit.ai: E-mail: research@whiterabbit.ai}

\begin{document} 
\maketitle

\begin{abstract} 
Mammography-based screening has helped reduce the breast cancer mortality rate, but has also been associated with potential harms due to low specificity, leading to unnecessary exams or procedures, and low sensitivity. 
Digital breast tomosynthesis (DBT) improves on conventional mammography by increasing both sensitivity and specificity and is becoming common in clinical settings. However, deep learning (DL) models have been developed mainly on conventional 2D full-field digital mammography (FFDM) or scanned film images.
Due to a lack of large annotated DBT datasets, it is difficult to train a model on DBT from scratch. In this work, we present methods to generalize a model trained on FFDM images to DBT images. In particular, we use average histogram matching (HM) and DL fine-tuning methods to generalize a FFDM model to the 2D maximum intensity projection (MIP) of DBT images. In the proposed approach, the differences between the FFDM and DBT domains are reduced via HM and then the base model, which was trained on abundant FFDM images, is fine-tuned. When evaluating on image patches extracted around identified findings, we are able to achieve similar areas under the receiver operating characteristic curve (ROC AUC) of $\sim 0.9$ for FFDM and $\sim 0.85$ for MIP images, as compared to a ROC AUC of $\sim 0.75$ when tested directly on MIP images. 

\end{abstract}



\section{PURPOSE}
\label{sec:purpose}  

Breast cancer is the most commonly diagnosed cancer and the second leading cancer-related cause of death among women in the United States \cite{siegel:2019}. Although mammography-based screening has been shown to reduce breast cancer mortality\cite{pmid30411328}, it has also been associated with physical and psychological harms caused by false positives and unnecessary biopsies \cite{smith:2004,tabar:2015, webb:2014}. To address these concerns, many clinics have started to switch their screening programs from 2D full-field digital mammography (FFDM) to 3D digital breast tomosynthesis (DBT)\cite{richman:2019}, which has been shown to increase the sensitivity of breast cancer screening\cite{hooley:2017, yueh:2019} and reduce false positives\cite{friedewald_breast_2014}. 

Deep learning (DL) using convolutional neural networks has previously been used to aid in the evaluation of screening mammography to enhance the specificity of malignancy prediction, particularly for FFDM exams\cite{wentao:2017, nyu:2019}. Several challenges exist, however, in translating these successes to DBT exams.
First, in general, the performance of DL models scales with the availability of labeled data but as a result of DBT being only more recently adopted, most large-scale mammography datasets consist mainly of FFDM exams. Second, the 3D volumes of DBT exams can be quite large (e.g., 2457$\times$1996$\times$70~pixels). This can lead to computational difficulties as well as training issues related to the curse of dimensionality, which are exacerbated by the low prevalence of cancer training samples and the often small finding sizes associated with the early detection.


This study focuses on methods to adapt DL malignancy models originally developed for FFDM exams to DBT exams in the case where the amount of available DBT data is quite limited. In order to overcome the large size of 3D DBT images, we instead consider the maximum intensity projection (MIP) of these 3D volumes. Several methods of adapting a model trained on patches of FFDM images to patches of MIP images are evaluated and compared. The impact of histogram matching on reducing domain shift and simplifying the adaptation problem is also considered. 


\section{METHODS}

\subsection{Dataset}
\label{sec:dataset}

The data was collected from a large academic medical center located in the mid-western region of the United States between 2008 and 2017. This study was approved by the internal institutional review board of the university from which the data was collected. Informed consent was waived for this retrospective study. The data consists of a large set of FFDM exams and a smaller set of DBT exams. The exams were interpreted by one of 11 radiologists with breast imaging experience ranging from 2 to 30 years. Radiologist assessments and pathology outcomes were extracted from the mammography reporting software of the site (Magview v7.1, Magview, Burtonsville, Maryland).

Patients were randomly split into training (`train'), validation (`val') and testing (`test') sets with a 80:10:10 ratio. Since the split was performed at the patient level, no patient had images in more than one of the above sets. This split was shared by both the FFDM and DBT datasets. All training and hyperparameter searches were performed on the training and validation sets. Performance on the test set was evaluated only once all model selection, training, and fine-tuning had been carried out.

Images were categorized as one of four classes: (1) normal, no notable findings were identified by the radiologist; (2) benign, all notable findings were determined to be benign by the radiologist or by biopsy, (3) high-risk, a biopsy determined a finding to contain tissue types likely to develop into cancer, and (4) malignant, a biopsy determined a finding to contain malignant tissue types.
We combine normal and benign labels into the negative class and combine the high risk and malignant labels into the positive class.
All malignant and high risk images had exactly one radiologist annotation, indicating the location of the biopsied finding. These annotations were made during the course of the standard clinical care for the patient. Benign samples have zero or one annotation. A detailed distribution of the data across the different classes can be found in Table~\ref{tab:data}.

\begin{table}[bhp]
\caption{Detailed statistics of the collected FFDM and DBT training (train), validation (val), and testing (test) datasets. The numbers of patients, exams, and images for each set are given, as well as the distribution of malignancy by image.} 
\label{tab:data}
\begin{center}       
\begin{tabular}{P{1.6cm} P{2.4cm} P{2.2cm} P{2.2cm}} 
\toprule
\multicolumn{4}{c}{\textbf{FFDM}}\\
~ & \textbf{Train} & \textbf{Val} & \textbf{Test} \\
\hline

Patients & 49965 & 6239 & 6213 \\
Exams & 158650 & 19933 & 19618 \\
Images & 664234 & 83920 & 82296 \\
\hline
Normal & 606080 (91.3\%) & 76092 (90.7\%) & 75073 (91.2\%) \\
Benign & 56660 (8.5\%) & 7631 (9.1\%) & 7023 (8.5\%) \\
High Risk & 404 (0.1\%) & 41 ~~(0.1\%) & 42 ~~(0.1\%) \\
Malignant & 1090 (0.2\%) & 156 (0.2\%) & 158 (0.2\%) \\
\midrule
\multicolumn{4}{c}{\textbf{DBT}}\\
~ & \textbf{Train} & \textbf{Val} & \textbf{Test} \\
\hline
Patients & 10684 & 1399 & 1357 \\
Exams & 14828 & 1944 & 1855 \\
Images & 54380 & 7140 & 6791 \\
\hline
Normal & 48006 (88.3\%) & 6171 (86.4\%) & 6058 (89.2\%) \\
Benign & 6175 (11.4\%) & 939 (13.2\%) & 689 (10.2\%) \\
High Risk & 86 ~~(0.2\%) & 13 ~~(0.2\%) & 15 ~~(0.2\%) \\
Malignant & 113 (0.2\%) & 17 ~~(0.2\%) & 29 ~~(0.4\%) \\
\bottomrule
\end{tabular}
\end{center}
\end{table}

\subsection{Patch Model}
\label{sec:patch_model}

The DL model is a ResNet\cite{resnet:2016} based model with 29 layers and approximately 6 million parameters. It accepts a 512x512 image patch from an FFDM or MIP image and predicts the probability that the patch contains a malignant or high risk finding. 

\begin{figure}
\centering
\includegraphics[width=\textwidth]{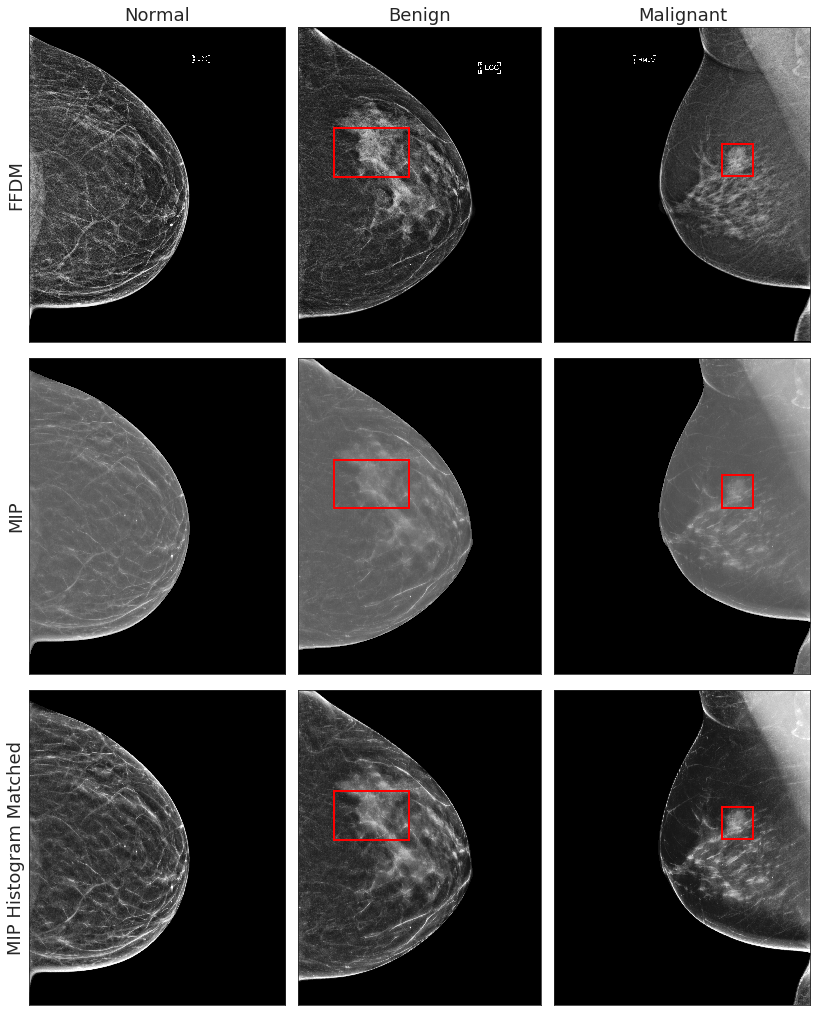}
\caption[sample]{\label{fig:samples}Sample images from each domain for different malignancy classes - Normal, Benign and Malignant. The red box indicates the location of a radiologist-annotated finding.}
\end{figure}

The original images had 4096$\times$3328 or 3328$\times$2560 pixels for FFDM images or 2457$\times$1996 or 2457$\times$1890 pixels for the MIP images. Example FFDM and MIP images can be seen in Figure~\ref{fig:samples}. To obtain the input to the model, an initial patch of 1024$\times$1024 pixels is extracted from the image and downsampled by bilinear interpolation to 512$\times$512 pixels, yielding a patch at half the resolution of the original image. The resulting patch covers 7.7-12.3\% of the area of the original image. 

For samples with annotations indicating the finding location, patches are centered at the center of the annotations. For samples without annotations, the breast is segmented using a pre-chosen threshold and a patch is selected centered at a randomly chosen pixel within the breast. Patches are sampled such that they are always fully contained within the image and may be translated to satisfy this criterion.

\subsection{FFDM Training}
The base model is trained on the FFDM data, with a uniform sampling of two classes (equal probability of sampling a positive or negative class sample). Images were augmented during training with random horizontal and vertical flipping, additive Gaussian white noise with a standard deviation of 1.0, random translation drawn from an Gaussian distribution with a standard deviation of 20 pixels, and random rotation drawn from an uniform distribution from -30 to +30 degrees.

The model is trained to minimize a cross entropy loss function using the Adam optimizer\cite{kingma_adam:_2015} with an initial learning rate of $5 \times 10^{-5}$ and a weight decay of $5 \times 10^{-4}$. An epoch is defined as 40000 samples shown to the model.
The model was trained for 100 epochs, and the model chosen for evaluation is the one that maximized the area under the receiver operating characteristic curve (ROC AUC) on the validation set.

\subsection{Domain Adaptation}

\begin{wrapfigure}[16]{r}{0.5\textwidth}
\vspace{-15mm}
\centering
\includegraphics[width=0.5\textwidth]{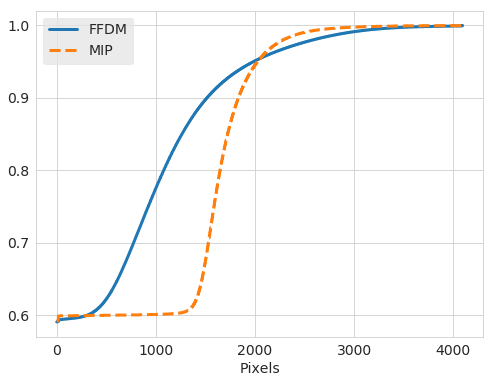}
\caption[cdf]{ \label{fig:cdf} Average cumulative histograms for FFDM and MIP images. The intensity values have been scaled so that they range from 0 to 4095 for both image types.}
\vspace{-25mm}
\end{wrapfigure}%

In domain adaptation, a model trained on one domain is adapted to another domain for which there exists far less data. Previous work has shown that deep neural networks often learn task and domain agnostic features, particularly in the earlier layers of the network\cite{yosinski:2014}. When the domains and tasks are similar, larger portions of the network may be reused.

Here, we explore the use of histogram matching to reduce the domain shift between the FFDM and DBT domains. The FFDM patch model is adapted to DBT exams, both with and without histogram matching, using two different fine-tuning methods.

\subsubsection{Histogram Matching}
\label{sec:histmatch}

A non-linear transformation is used to transform the cumulative histogram (c.d.f.) from one domain to the average c.d.f. of another domain\cite{gonzalez:2017}, referred to as histogram matching (HM). In particular, HM is employed to transform MIP images to better match the FFDM images originally used to train the model.

\setcounter{algocf}{1}
\begin{center}
\begin{minipage}[t]{0.75\textwidth}
 \begin{algorithm}[H]
    \SetAlgoLined
    \KwIn{Source image $X_S \in \left[0, K-1 \right]^N$, \\ 
    Source c.d.f. $F_S \in \mathbb{N}_0^K$, \\ 
    Reference c.d.f. $F_R \in \mathbb{N}_0^L$}
    \KwOut{Histogram-matched image $X_S' \in \left[0, L-1\right]^N$}
    \For{$i$ in 0 to N-1}{
    $p_S = X_S[i]$ \\
    $p_S' = F_R^{-1}\left(F_S(p_S)\right)$\\
    $X_S'[i] = p_S'$
    }
    \Return $X_S'$
    \caption*{\textbf{Histogram matching} \\ The algorithm describes the procedure of histogram matching images from two domains. Here, $X[i]$ represents the $i$-th pixel value of the image $X$. The inverse mapping to a pixel value in the reference domain is performed by linear interpolation.} \label{alg:histogram_matching}
\end{algorithm}
\end{minipage}
\end{center}

The procedure for HM is outlined in Algorithm~\ref{alg:histogram_matching} and given in greater detail as follows. Let $F_S$ be the c.d.f. of the source image, whose intensity distribution is to be updated, and let $F_R$ be the c.d.f. of the reference domain, whose intensity distribution we hope to match. Let $p_S \in \left[0, K-1 \right]$ be a pixel value for the source image and $p_R \in \left[0, L-1\right]$ be a pixel value in the reference domain such that $F_S(p_S) = F_R(p_R)$. 
Then, our transformed image will have the pixel value $p_S' = p_R =F_R^{-1}\left(F_S(p_S)\right)$, where the inverse mapping is calculated via linear interpolation.

The average c.d.f. of the FFDM data was calculated over 1200 randomly chosen training samples, comprised of equal amounts of the normal, benign and malignant classes. Similarly, the average c.d.f. of the MIP data was calculated over 600 randomly chosen training samples, comprised of equal amounts of the normal, benign and malignant classes. The histograms for the FFDM and MIP images can be seen in Figure \ref{fig:cdf}. The application of histogram matching can be qualitatively visualized in Figure \ref{fig:samples}.

\subsubsection{Fine-tuning}
\label{sec:finetune}
Two methods were used to fine-tune the base model trained on FFDM images for use with the original or histogram-matched MIP images. For the first approach, only the last fully connected layer of the model was re-trained. This is referred to as the conventional fine-tuning approach. For the second approach, a version of the SpotTune algorithm was implemented\cite{guo2019spottune}.
SpotTune is an adaptive fine-tuning approach that finds the optimal fine-tuning policy (which layers to fine-tune) per instance of target data.

The underlying idea behind SpotTune is that different training samples from the target domain require fine-tuning updates to different sets of layers in pre-trained network. The SpotTune training procedure involves predicting, for each training input, the specific layers to be fine-tuned and layers to be kept frozen. This input-dependent fine-tuning approach enables targeting layers per input instance and leads to better accuracy~\cite{guo2019spottune}. We refer readers to the original paper\cite{guo2019spottune} for further details of SpotTune.

The fine-tuned model used the same data augmentations as the original FFDM model. The model is fine-tuned using cross entropy loss and Adam optimizer with a learning rate of $5 \times 10^{-5}$ and a weight decay of $1 \times 10^{-4}$. The model chosen is the one that maximized the validation ROC AUC.

\section{RESULTS}
\label{sec:results}

The performance of all models is measured on the test datasets using the area under the receiver operating characteristic curve (ROC AUC). On the test data, we extract patches in the same way as explained in Section \ref{sec:patch_model}.  Since this is random, we average the results over three random seeds. The standard deviation of results is used as an error estimate. A summary of all the results can be found in Table \ref{tab:results}.

\begin{table}[htbp]
\caption{Performance of the models for different domains. Results are shown on a test set for which both FFDM and DBT/MIP images are available. MIP with HM refers to MIP images pre-processed to look more like FFDM images. Errors shown here refer to the standard deviation over 3 independent realizations of patch extraction.} 
\label{tab:results}
\begin{center}       
\begin{tabular}{P{3.5cm} P{3.5cm} P{3.5cm} P{3.5cm}} 
\hline
\hline
\textbf{Training Data} & \textbf{Testing Data} & \textbf{Procedure} & \textbf{ROC AUC}  \\
\hline
\hline
FFDM & FFDM & Train from scratch & $0.909 \pm 0.001$  \\
FFDM & MIP & Test only & $0.751 \pm 0.001$  \\
FFDM & MIP & Fine-tune & $0.759 \pm 0.003 $  \\
FFDM & MIP & SpotTune~\cite{guo2019spottune} & $0.825 \pm 0.002$  \\
\hline
FFDM & MIP with HM & Test only & $\mathbf{0.847 \pm 0.001}$  \\
FFDM & MIP with HM & Fine-tune & $0.837 \pm 0.001 $  \\
FFDM & MIP with HM & SpotTune~\cite{guo2019spottune} & $0.830 \pm 0.002 $  \\
\hline 
\hline
\end{tabular}
\end{center}
\end{table}

The base FFDM patch model has a ROC AUC of 0.909 on the FFDM images. For MIP images, the performance of the base model drops to a ROC AUC of 0.751. If the MIP images are pre-processed using the average histogram matching method, the ROC AUC goes up to 0.847. This shows that our simple fixed non-linear transformation via histogram matching reduces the domain shift considerably.

In order to improve performance further, we apply fine-tuning and SpotTune\cite{guo2019spottune}, both with and without HM. Fine-tuning on the limited number of regular MIP images does not show any improvement in performance; however, SpotTune leads to a ROC AUC of 0.825. When fine-tuned on MIP images with HM, fine-tuning improves ROC AUC from 0.759 to 0.837; however, SpotTune leads only to a minor improvement in ROC AUC from 0.825 to 0.830.
Overall, we find that the simple strategy of only pre-processing via histogram matching leads to the best ROC AUC of 0.847 on MIP images.

\begin{wrapfigure}{r}{0.5\textwidth}
\vspace{-5mm}
\centering
\includegraphics[width=0.5\textwidth]{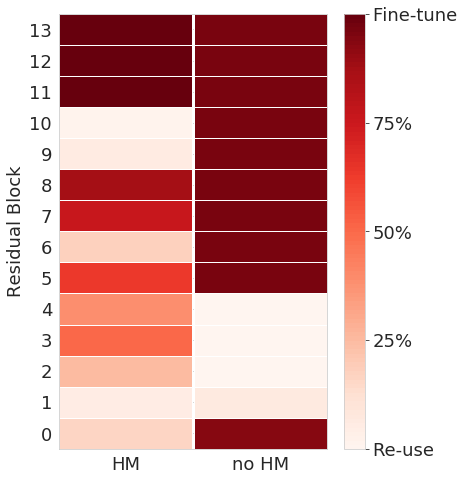}
\caption[spottune]{\label{fig:spottune}Visualization of SpotTune policies to re-use or fine-tune a residual block for MIP images with and without HM.}
\vspace{-10mm}
\end{wrapfigure}

SpotTune learns a policy per sample for selecting a ResNet block for fine-tuning. For two cases, with and without HM, the probability of fine-tuning different ResNet blocks can be seen in Figure \ref{fig:spottune}. With HM, the relative gap in performance of SpotTune vs. Test-only is small, perhaps due to the limited number of ResNet blocks that are modified by the SpotTune algorithm.  Without HM, the gap is large and a significant portion of the blocks are updated.

This observation also offers insight into the effectiveness of conventional fine-tuning with and without HM. Without HM, fine-tuning the last layer of the network is unable to improve performance as more extensive changes to the network are needed as indicated by the large number of ResNet blocks changed by the SpotTune algorithm. With HM, the performance of SpotTune and conventional fine-tuning are more similar as only the late layers need to be extensively modified.


\section{NEW OR BREAKTHROUGH WORK TO BE PRESENTED}
We present our work on the adaptation of a patch-level deep learning malignancy model from FFDM to DBT exams. The original model was trained and evaluated on a large set of the FFDM images and the model was adapted using methods requiring few DBT exams. In particular, by incorporating histogram level changes in image features, we can achieve good classification performance without additional training.

A prior study\cite{mendel:2019} considered the use of transfer learning for malignancy classification for FFDM and DBT exams. However, the study examine the transfer learning for a model that was not initially trained on medical images. It also mainly focused on malignant and benign patches (4$\times$ smaller than our patches) and employed a much smaller dataset.
A very recent study\cite{lotter_robust_2019} considered domain adaptation from FFDM to DBT images, but relied on a large multi-site dataset of DBT images. It is unclear how effective that approach would be if the DBT dataset were smaller.

\section{CONCLUSIONS}

A deep learning malignancy model was trained to identify high risk or malignant findings in patches of full-field digital mammography (FFDM) images. Several strategies were evaluated to adapt this model to the maximum intensity projections (MIP) of digital breast tomosynthesis (DBT) exams. 
The effectiveness of each domain adaptation approach depended strongly on the amount of domain shift and the amount of available training data. Without HM, the amount of domain shift was large and SpotTune was the most effective even with limited training data. However, following HM, the domain shift was much smaller and the simplest approach proved best. Histogram match can, therefore, be an effective strategy for domain adaptation when the amount of available training data in the target domain is limited.
This approach is simple and intuitive and can be easily adapted to other problems in medical imaging where obtaining a large amount of labeled data for every image modality is difficult.

\section{FUTURE WORK}

In this study, we focus exclusively on domain adaptation for patch-level models. Thus, future work includes using these adaptation techniques and patch models to train whole-image models, which provide a malignancy probability for the entire image and, eventually, the entire examination. Another interesting approach is learning the cyclic transformation via a CycleGAN \cite{DBLP:journals/corr/ZhuPIE17}. This generative model can learn how to transform a 3D DBT image into a 2D image from the same distribution as the original FFDM images (and vice versa). By learning this conditional distribution, we could synthesize 2D images that maintain the important, subtle features that may have been lost by using maximum projection.

\section{DISCLOSURES}
This work has not been submitted to any journal or conference for publication or presentation considerations.

This work was supported by Whiterabbit AI, Inc. The following authors are employed by and/or have a financial interest in Whiterabbit: Sadanand Singh, Thomas Paul Matthews, Meet Shah, Brent Mombourquette, Trevor Tsue, Aaron Long, Stefano Pedemonte, and Jason Su.

\appendix    
\bibliographystyle{spiebib}
\bibliography{report}

\end{document}